\documentclass{article}

\usepackage{PRIMEarxiv}

\usepackage[utf8]{inputenc} 
\usepackage[T1]{fontenc}    
\usepackage{hyperref}       
\usepackage{url}            
\usepackage{booktabs}       
\usepackage{amsfonts}       
\usepackage{nicefrac}       
\usepackage{microtype}      
\usepackage{lipsum}
\usepackage{fancyhdr}       
\usepackage{graphicx}       
\graphicspath{{media/}}     

\title{Measuring How LLMs Internalize Human Psychological Concepts: A preliminary analysis
\thanks{\textit{\underline{Citation}}: 
\textbf{Authors. Title. Pages.... DOI:000000/11111.}} 
}

\author{
  Hiro Taiyo Hamada, Ippei Fujisawa \\
  R\&D Department \\
  Araya inc.  \\
  Tokyo, Japan\\
  \texttt{\{hamada\_h, fujisawa\}@araya.org} \\
  \And
  Genji Kawakita \\
  Department of Computational Neuroscience \\
  Imperial College London \\
  London, United Kingdom\\
  \texttt{\{g.kawakita22\}@imperial.ac.uk} \\
  \And
  Yuki Yamada \\
  Kyusyu University \\
  Kyusyu University \\
  Fukuoka, Japan \\
  \texttt{email@email} \\
}

\begin{document}
\maketitle

\begin{abstract}
Large Language Models (LLMs) such as ChatGPT have shown remarkable abilities in producing human-like text. However, it is unclear how accurately these models internalize concepts that shape human thought and behavior. Here, we developed a quantitative framework to assess concept alignment between LLMs and human psychological dimensions using 43 standardized psychological questionnaires, selected for their established validity in measuring distinct psychological constructs. Our method evaluates how accurately language models reconstruct and classify questionnaire items through pairwise similarity analysis. We compared resulting cluster structures with the original categorical labels using hierarchical clustering. A GPT-4 model achieved superior classification accuracy (66.2\%), significantly outperforming GPT-3.5 (55.9\%) and BERT (48.1\%), all exceeding random baseline performance (31.9\%). We also demonstrated that the estimated semantic similarity from GPT-4 is associated with Pearson's correlation coefficients of human responses in multiple psychological questionnaires. This framework provides a novel approach to evaluate the alignment of the human-LLM concept and identify potential representational biases. Our findings demonstrate that modern LLMs can approximate human psychological constructs with measurable accuracy, offering insights for developing more interpretable AI systems.
\end{abstract}


\section{Introduction}

AI systems rapidly accelerate research and development. Applications include automating drug discovery experiments \cite{BoikoChemical2023}, exploring medical applications \cite{SinghalLLMClinical2023}, and developing alternative sampling and intervention methods in psychology \cite{DemszkyPsychol2024}. Psychological research examines whether Large Language Models (LLMs) can diagnose personality traits \cite{YangPsyCoT2023}, substitute participant responses \cite{DillionParticipants2024}, and mimic fictional character responses. Some studies also showed that LLMs imitate human sensory modalities and cognitive biases \cite{KawakitaColor2024, taleb2024}. These studies suggest that LLMs like GPT-4, trained from vast text data, may learn relationships between psychological concepts and related human linguistic patterns.

Measuring human concept internalization in AI systems is important since AI systems may have different concept representations that humans have. It is also known that even  representations of human concepts may have different semantics in different cultures \cite{JacksonEmotion2019}. By understanding different systems like human cultures, we may have better communications with AI systems. This human-AI alignment at the concept level is called concept alignment \cite{RaneConceptAlign2024}. Analyzing them gives us clues on how AI systems internalize concept representations. However, it remains unclear whether language models internalize such concepts. 

Here, we proposed a novel method to study whether language models, including GPT-4, can internalize conceptual category classifications from psychological questionnaire items. We first compared model classification performance by calculating similarities between items from 43 psychological questionnaires using language models like GPT-3.5 and GPT-4, and measured agreement between similarity-based classifications and original construct labels. We found that GPT-4 showed the highest classification score among other language models while the OpenAI embedding model was comparable. Then, we checked if continuous and discrete evaluations with prompts have different classification performances. The results did not show statistical difference in classification accuracy. We also checked if there is ordering effect of questionnaire items on classification performance. Some questionnaires showed different classification performances. We finally showed that estimated semantic similarity from multiple questionnaires is linked with Pearson's correlation coefficients of human responses to the questionnaires.

Our findings suggest that LLMs retain relationships between psychological concepts. Our framework provides a new tool to measure the level of concept alignment of AI systems.

\section{Related Works}
\subsection{Replicating Human Knowledge and Personality Traits with Language Models}
Due to the emergence of LLMs, multiple studies explored their ability to replicate human knowledge and personality traits across various domains \cite{DemszkyPsychol2024, SatoriPsySci2023, YangPsyCoT2023}. 

In psychology, studies showed reflection of population responses in LLMs. Some studies attempt to imitate human groups and responses of fictional characters \cite{li2023chatharuhi, ArgyleHumanSamples2023}. \cite{ArgyleHumanSamples2023} used GPT-3 to answer demographic questions using subject's demographic information, and found statistically significant correlations between human samples and GPT-3's responses \cite{ArgyleHumanSamples2023}. \cite{li2023chatharuhi} trained models on anime character dialogues to complete personality questionnaires\cite{li2023chatharuhi}. \cite{CulterLexical2023} found association between between human Big Five responses and adjective similarity ratings of the language models \cite{CulterLexical2023}. Other studies also examine whether learned sensory relationships of LLMs reflect human cognitive structures \cite{KawakitaColor2024, ChenCBias2024}. \cite{KawakitaColor2024} showed color similarity structure of neuro-typical subjects is more aligned with that of GPT-4 than that of GPT-3.5 \cite{KawakitaColor2024}. 

These studies analyze LLMs which were trained from vast text and image data using psychological testing methods. The correspondence between human and LLM responses suggests the possibility of obtaining equivalent information through LLMs without direct human experiments. This framework aims to understand the responses of AI systems in cognitive tasks, also known as machine psychology \cite{HagendorffMachinePsy2024}.

In this work, we focus on a framework to measure how LLMs internalize psychological dimensions from standardized questionnaires.

\section{METHOD}
\subsection{Database}
From the psychological questionnaire database "Psychological Scales" ("https://scales.arabpsychology.com/"), we extracted 43 English psychological questionnaires contain 30 or fewer text-based items. Each questionnaire is developed based on psychological constructs such as "curiosity," "self-efficacy," and "anxiety," and includes between 2 and 6 sub-categories labeled by experts.

\subsection{Pairwise similarity analysis for classification}
\label{pairclassification}
\begin{figure}[h]
\begin{center}
    \includegraphics[width=1.0\textwidth]{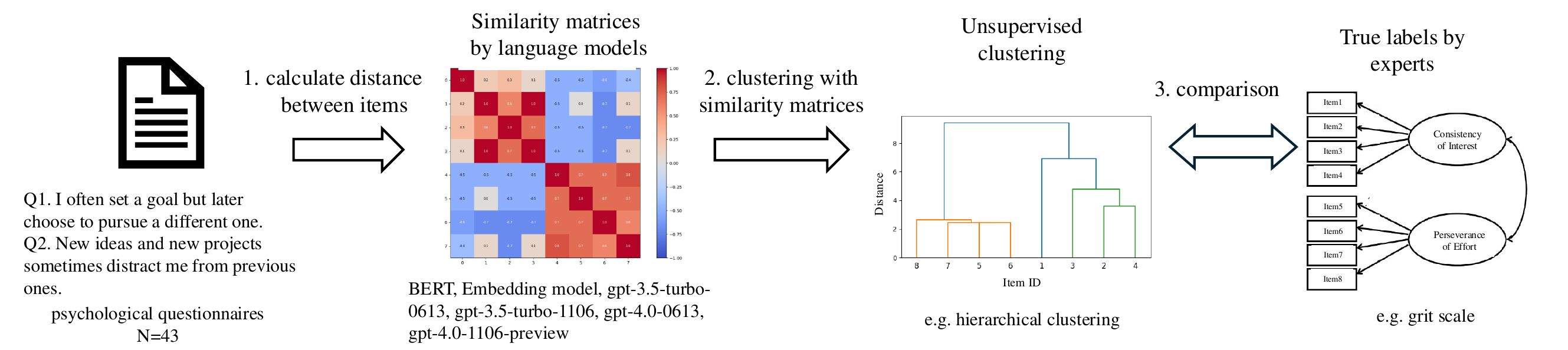}
    \caption{Classification with pairwise similarity analysis. 1. A distance between two items from each psychological questionnaire is calculated with language models from -1 to 1. 2. Each similarity matrix was classified with hierarchical clustering with predetermined number of categories for each questionnaire. 3. The labels with unsupervised clustering were compared with true labels with to calculate classification accuracy.}
    \label{figure1_realign}
\end{center}
\end{figure}

We compare the classification performance of six language models including bert-base-uncased, Open AI Embedding, gpt-3.5-turbo-0613, gpt-3.5-turbo-0613, gpt-4-turbo-0613 
, and gpt-4-turbo-1106-preview on questionnaire items (Table.~\ref{table1}). For BERT and an embedding model, we embed all questions from each questionnaire and calculate cosine similarity between items as a similarity matrix (Fig.~\ref{figure1_realign}). For GPT-3.5 and GPT-4, we provide prompts instructing them to respond within the [-1, 1] range based on semantic similarity between questionnaire items (Table. ~\ref{table2}). Here, 1 indicates that two items mean identical while -1 indicates that two items mean opposite. The 0 score indicates no similarity between the items. The responses form similarity matrices for each questionnaire.

Finally, we apply hierarchical clustering to the generated similarity matrices, using the predetermined number of categories for each questionnaire. For text classification metrics, we use the classification accuracy between classified items and their true categories, along with the adjusted rand index (ARI).

\begin{table}[t]
\begin{center}
\begin{tabular}{|c|l|p{8cm}|}
\hline
\textbf{ID} & \textbf{Questionnaire Name} & \textbf{Abbreviation}  \\
\hline
1        &bert-base-uncased (BERT) & Parameters: 110M \\
\hline
2        &OpenAI embedding & An embedding model provided by OpenAI "text-embedding-ada-002" \\
\hline
3        &gpt-3.5-turbo-0613 (gpt-3.5-0613) & OpenAI API specifications: Parameters: Not disclosed,
Context window: 4,096 tokens, Training data: Up to September 2021\\
\hline
4        &gpt-3.5-turbo-1106 (pt-3.5-1106) & OpenAI API specifications: Parameters: Not disclosed,
Context window: 16,385 tokens, Training data: Up to September 2021 \\
\hline
5        &gpt-4-turbo-0613 (gpt-4-0613) & OpenAI API specifications: Parameters: Not disclosed,
Context window: 8,192 tokens, Training data: Up to September 2021 \\
\hline
6        &gpt-4-turbo-1106-preview (gpt-4-1106) & OpenAI API specifications: Parameters: Not disclosed,
Context window: 128,000 tokens, Training data: Up to April 2023 \\
\hline
7        &gpt-4-turbo-1106-preview (gpt-4-1106) & OpenAI API specifications: Parameters: Not disclosed,
Context window: 128,000 tokens, Training data: Up to April 2023 \\
\hline
8       &gpt-4-turbo-1106-preview (gpt-4-1106) & OpenAI API specifications: Parameters: Not disclosed,
Context window: 128,000 tokens, Training data: Up to April 2023 \\
\hline
\end{tabular}
\caption{Overview of language models and their specifications}
\label{table1}
\end{center}
\end{table}

\subsection{Pairwise similarity analysis across multiple questionnaires}
\label{pairwiseacross}
We apply pairwise similarity analysis to describe semantic relationship across multiple psychological questionnaires. Item-by-item similarity scores across different psychological questionnaires are calculated. In this study, we check whether Pearson's correlation results of 8 dimensions are used from psychological questionnaires \cite{DinovoSelfRegu2011}. The questionnaires include positive affect (PA) and negative affect (NA) from the Positive and Negative Affect Schedule (PANAS) \cite{WatsonPANAS1995}, behavioral inhibition (BIS) and behavioral activation (BAS) from the Behavioral Inhibition System and Behavioral Activation System Scales (BIS/BAS Scales) \cite{CarverBISBAS1994}, general distress and anhedonia from the Mood and Anxiety Symptom Questionnaire (MASQ) \cite{WatsonMASQ1995}, and self-deception positivity (SD) and impression management (IM) from the Balanced Inventory of Desirable Responding (BIDR) \cite{PaulhusBIDR2024}. 

We first calculated the mean and median pairwise similarity scores between items in multiple questionnaires (Fig.\ref{Sfigure3_realign}). Then, we constructed a similarity matrix and it is compared to the correlation matrix of the original study \cite{DinovoSelfRegu2011}.

\section{RESULTS}
In the following analyses, we performed pairwise similarity analysis for the classification of language models within psychological questionnaires and evaluating semantic relationships across the questionnaires. First, we assessed performance differences between language models through questionnaire item classification tasks. Next, we compared performance variations across different prompting approaches. Then, we examined how item order affects classification performance. We finally checked if there is an association between estimated semantic similarity from questionnaires with LLMs and correlation coefficients of human responses on these questionnaires.

\subsection{Reconstruction of psychological constructs via pairwise similarity}
\begin{figure}[ht]
\begin{center}
    \includegraphics[width=1.0\textwidth]{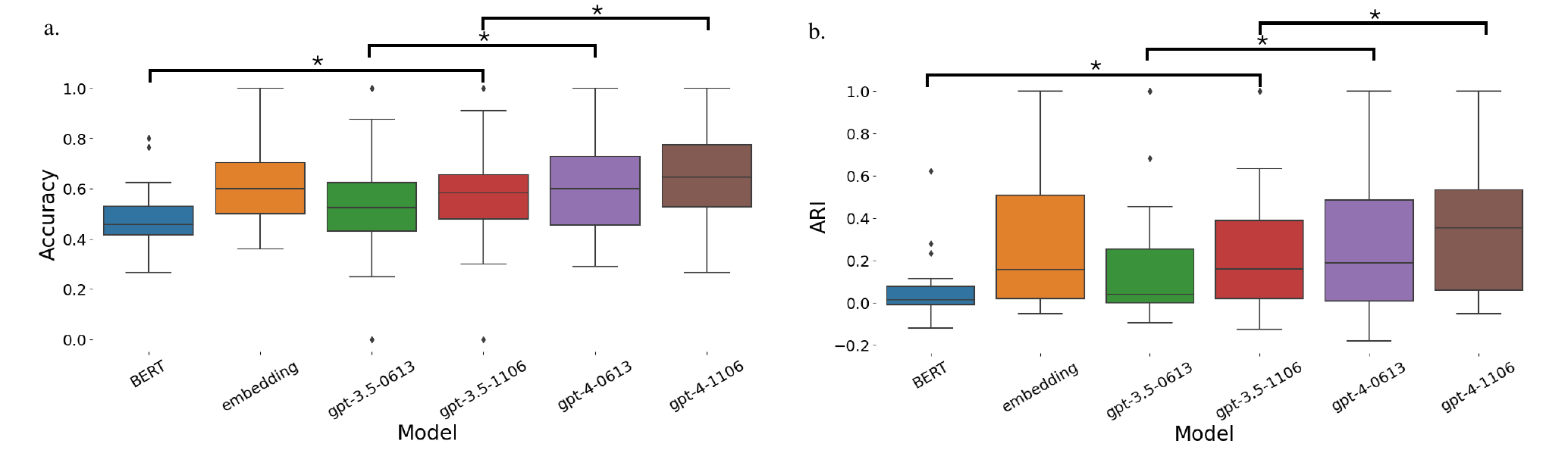}
    \caption{Classification Performance with language models. a. Mean accuracies of the classification performance were calculated with six language models including BERT, OpenAI embedding model, gpt-3.5-0613, gpt-3.5-1106, gpt-4-0613, and gpt4-1106. b. Mean ARIs of the classification performance were similarly calculated with the six language models. * indicates statistical significance (p $<$ 0.05).}
    \label{figure2_realign}
\end{center}
\end{figure}

First, we performed a classification task among multiple language models including bert-base-uncased, Open AI Embedding, gpt-3.5-turbo-0613, gpt-3.5-turbo-0613, gpt-4-turbo-0613 
, and gpt-4-turbo-1106-preview. We create similarity matrices from 43 standardized psychological questionnaires, and calculate the accuracy with questionnaire category labels from the questionnaires with statistical comparisons (paired Student's t-test) among 6 language models (see Table. ~\ref{table1}). Our results showed that GPT-4(1106) achieved the highest accuracy at 66.1\% and ARI, while BERT showed the lowest at 48.1\% above the random baseline 31.9\%. No statistical significance was found in mean classification accuracy and ARI between OpenAI embedding model, GPT-4(0613), and GPT-4(1106) (Fig.~\ref{figure2_realign}). 

Next, in order to check if the performance of gpt-4 models (0613/1106; accracy: 63.9\% and ARI: 0.341) is superior to that of gpt-3.5 models (0613/1106; accracy: 56.7\% and ARI: 0.210), we performed a statistical comparison between these models. Our analysis showed statistically significant higher classification performance with GPT-4 (0613/1106) (p $<$ 0.001, Fig.~\ref{Sfigure1_realign}).

Previous studies indicated that human-like prompts influence LLM performance \cite{taleb2024}. Discrete scales, a.k.a. Likert scales, are often used in psychological questionnaires so the scales may influence performance of the mean accuracy. We checked if the continuous and discrete prompts have different accuracy performances (Table. ~\ref{table2}, Fig.~\ref{Sfigure2_realign}). Our results did not show statistical significance between continuous and discrete prompts with the models gpt-4-0613 (p=0.45) and gpt-4-1106 (p=0.69).

\begin{figure}[ht]
\begin{center}
    \includegraphics[width=1.0\textwidth]{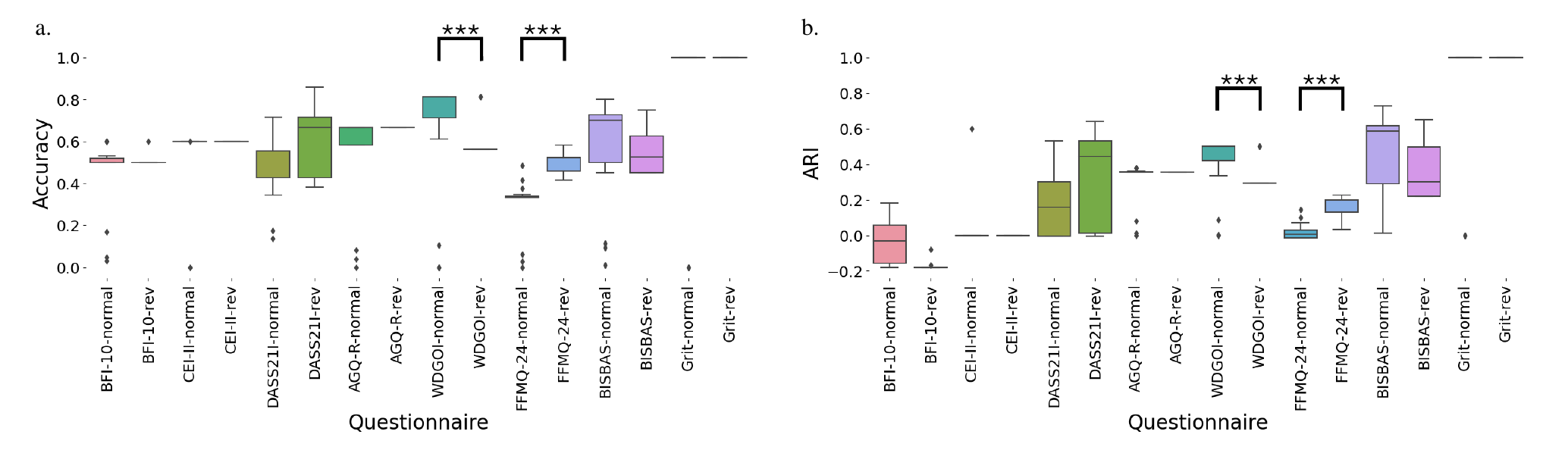}
    \caption{Ordering effect in the classification performance. a. Mean accuracies of the classification performance in different 8 questionnaires calculated with gpt4-1106. b. Mean ARIs of the classification performance in different 8 questionnaires calculated with gpt4-1106. The suffixes, -normal and -rev, indicate item-by-item calculation with normal and reverse orders, respectively (see the detail in the section \ref{pairclassification}). * indicates statistical significance (p $<$ 0.05).}
    \label{figure3_realign}
\end{center}
\end{figure}

\begin{figure}[ht]
\begin{center}
    \includegraphics[width=1.0\textwidth]{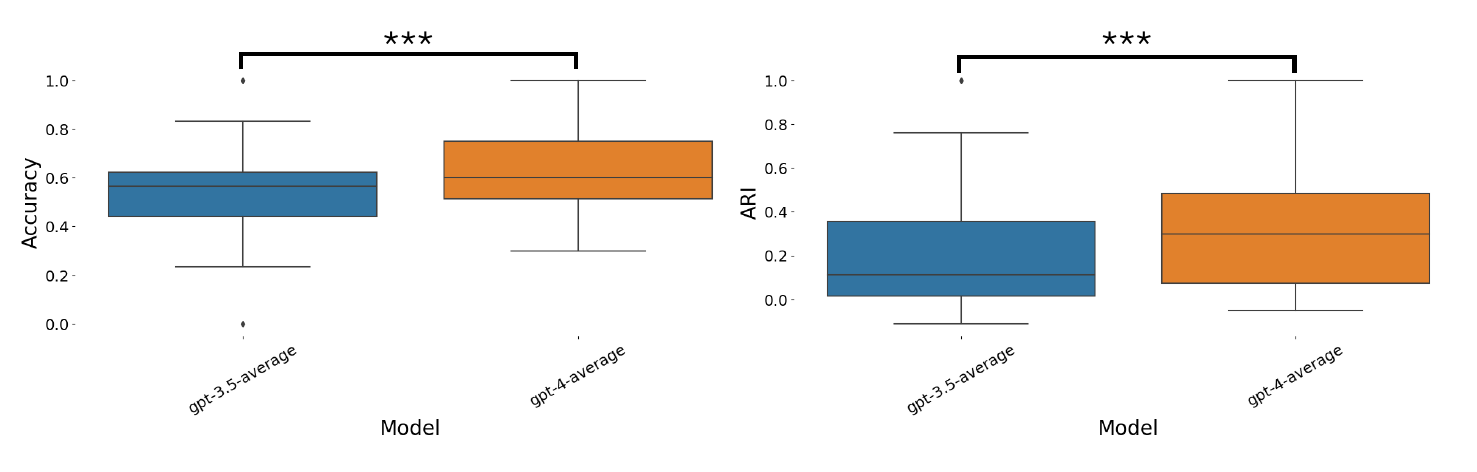}
    \caption{Comparison between gpt-3.5 and gpt-4 models. Mean accuracies and ARI of classification between gpt-3.5 models (gpt-3.5-0613/1106) and gpt-4 models (gpt-4-0613/1106) were compared. *** indicates p $<$ 0.001. }
    \label{Sfigure1_realign}
\end{center}
\end{figure}

\begin{figure}[ht]
\begin{center}
    \includegraphics[width=0.5\textwidth]{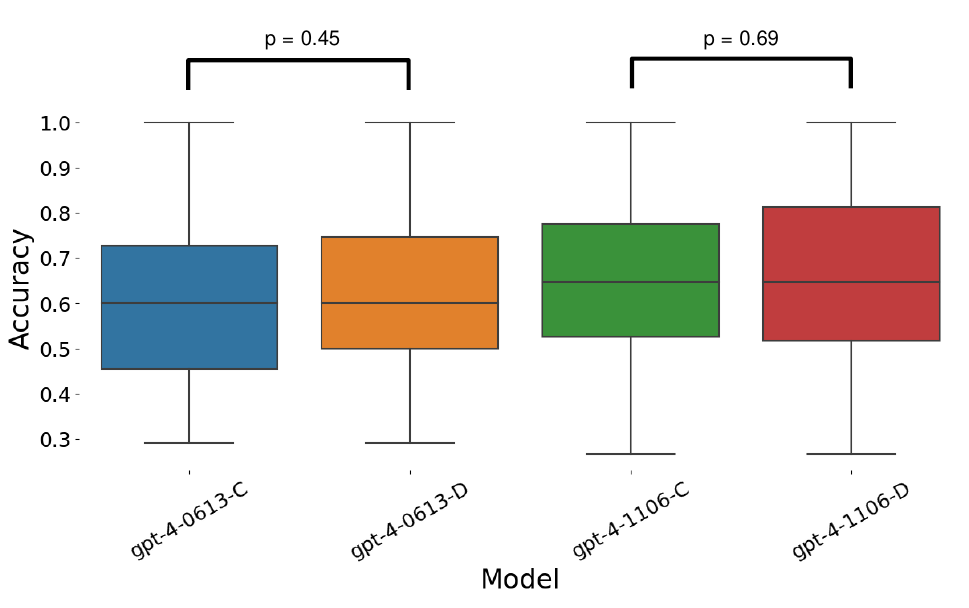}
    \caption{Comparison between continuous and discrete prompts. Accuracies between continuous and discrete prompts were compared with gpt-4-0613 and gpt-1106 (see Table. ~\ref{table2}). Labels, "-C" and "-D" indicate continous and discrete prompts, respectively.}
    \label{Sfigure2_realign}
\end{center}
\end{figure}

LLMs are sensitive to the ordering of the premises, so-called order effect \cite{ChenOrder2024}. We checked if there is an effect on this task with eight psychological questionnaires (Table. ~\ref{table3}). We compared mean accuracy with normal and reverse order of items in each questionnaire. Statistical significances were found in WDGOI and FFMQ-24 questionnaires (p $<$ 0.001, multiple comparisons with Bonferroni correction; Fig.~\ref{figure3_realign}). This result indicates that there is an ordering effect on this task.

\begin{table}[t]
\begin{center}
\begin{tabular}{|c|p{8cm}|}
\hline
\textbf{Category} & \textbf{Prompt}  \\
\hline
continuous &You are an expert of natural language. You will get two questions. You should answer the similarity score of two questions between -1 and 1 as a real number. A score close to 1 indicates high similarity, while a score closes to -1 suggests two questions have opposite meanings. A score around 0 implies two questions are neither similar nor opposite. If the similar You must not answer anything else neither add an explanation. \\
\hline
discrete (Likert scale)        &You are an expert of natural language. You will get two questions. You should answer the similarity score of two questions from 1,2,3,4,5,6,7,8, and 9. A score close to 9 indicates very similar, while a score closes to 1 suggests two questions are not very similar. A score around 5 implies two questions are neither similar nor opposite. If the similar You must not answer anything else neither add an explanation.\\
\hline
\end{tabular}
\caption{Descriptions of prompts}
\label{table2}
\end{center}
\end{table}

\begin{table}[t]
\begin{center}
\begin{tabular}{|c|l|p{2cm}|}
\hline
\textbf{ID} & \textbf{Questionnaire Name} & \textbf{The number of items}  \\
\hline
1        &The 10-Item Big Five Inventory (BFI-10) & 10 \\
\hline
2        &The Curiosity and Exploration Inventory-II (CEI-II) & 10 \\
\hline
3        &The Depression, Anxiety and Stress Scale - 21 Items (DASS-21)  & 21 \\
\hline
4        &Achievement Goal Questionnaire – Revised (AGQ-R) & 12 \\
\hline
5        &Work Domain Goal Orientation Instrument (WDGOI) & 16 \\
\hline
6        &Th 24 items version of Five Facet Mindfulness Questionnaire (FFMQ-24) & 24 \\
\hline
7        &Behavioral Inhibition System and Behavioral Activation System Scale (BIS/BAS) & 20 \\
\hline
8       &Grit Scale (Grit) & 8 \\
\hline
\end{tabular}
\caption{Psychological questionnaires for checking an ordering effect}
\label{table3}
\end{center}
\end{table}

Our findings suggest that language models internalize psychological categories, and different GPT models have different performance.

\begin{figure}[h]
\begin{center}
    \includegraphics[width=1.0\textwidth]{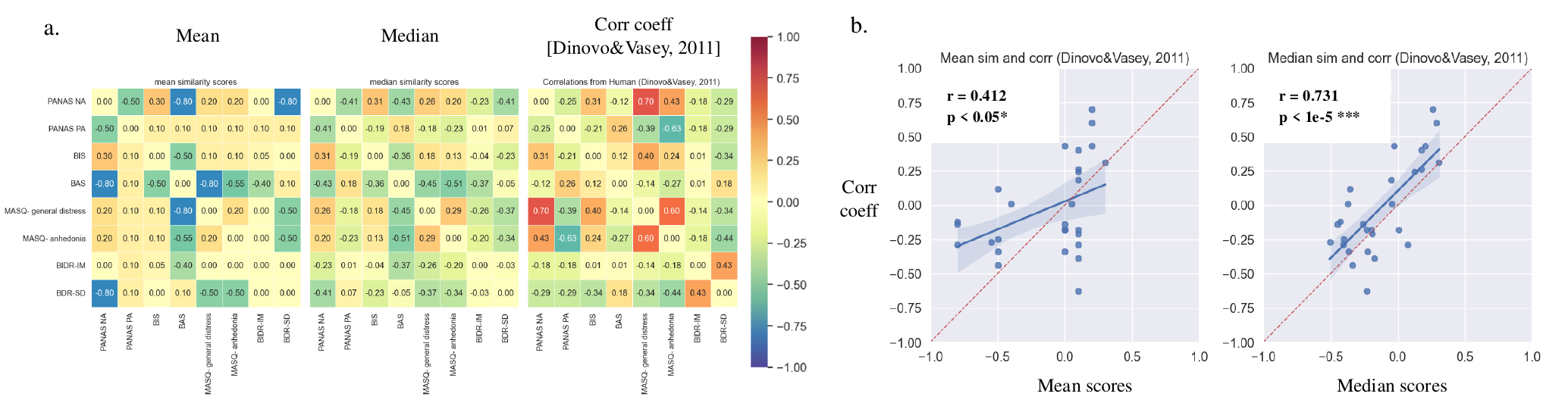}
    \caption{Association between estimated semantic similarity and human responses. a. The mean and median matrices of estimated semantic similarity were calculated with 8 psychological dimensions (see also the section \ref{pairwiseacross}). The corr coeff matrix indicates Pearson's correlation coefficients from human responses in psychological questionnaires \cite{DinovoSelfRegu2011}. b. Association of mean and median sores were calculated with the correlation coefficients from human responses (mean: r=0.412, p$<$0.05; median: r=0.731, p$<$1e-5). * and *** indicate p$<$0.05 and p$<$1e-5, respectively.}
    \label{figure4_realign}
\end{center}
\end{figure}

\begin{figure}[h]
\begin{center}
    \includegraphics[width=1\textwidth]{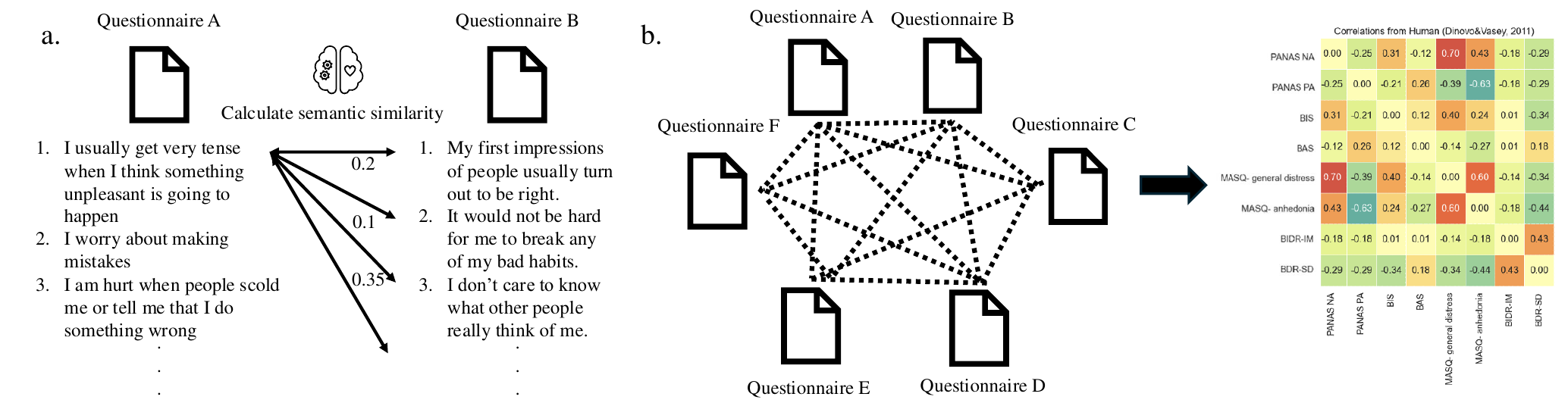}
    \caption{Pairwise similarity analysis for semantic relationship. a. Item-by-item semantic similarity is calculated between two questionnaires. b. Each mean or median  similarity score between items of two questionnaires is used as a link between questionnaires. A similarity matrix is constructed from these links across multiple questionnaires.}
    \label{Sfigure3_realign}
\end{center}
\end{figure}

\subsection{Semantic relationship between Psychological Constructs}
Next, we checked whether LLMs keep distance information between multiple psychological constructs. To check it, we used a previous result which asked subjects to answer multiple psychological questionnaires about the self-regulation temperament , such as effortful control with 6 other dimensions, and showed correlation between them (\cite{DinovoSelfRegu2011}; see also the section ~\ref{pairwiseacross} and Fig.~\ref{Sfigure3_realign}). We compared estimated distances between psychological dimensions from semantic similarity in the questionnaires with correlations from human responses in the original study. The results showed statistical significance with mean (r=0.412, p$<$0.05) and median scores (r=0.731, p$<$1e-5; Fig. \ref{figure4_realign}b). 

Our finding suggests that LLMs internalize semantic relationship between multiple psychological constructs.

\section{Discussion}
In this study, we examined whether language models internalize psychological dimension by category classifications of psychological questionnaire items based on constructs. In the first experiment, our finding showed that GPT-4 achieved 63.9\% average accuracy, surpassing GPT-3.5's 56.7\%. The embedding model demonstrated the performance comparable to that of GPT-4 with 63.9\% average accuracy. Second, our result showed the effect of order in the classification performance. Finally, we showed the association between estimated semantic similarity with a LLM and Pearson's correlation of human results from multiple psychological constructs. These results demonstrate the association between the output of language models and psychological dimensions, suggesting that they internalize information about psychological dimensions and related semantic patterns.

Our approach has two potential benefits. First, our proposed method can be a tool for evaluating concept-level alignment between human and LLMs. Human semantics in different cultures has universality and variations in their structures \cite{JacksonEmotion2019}. It is interesting to check whether LLMs embed such plural semantic structures. Second, semantic similarity between questionnaire items can used as a visualization tool for relationship between psychological concepts. While relationships between psychological concepts are hidden, language models allow us to visualize semantic relationship learned from vast texts in the world. We plan to validate with our proposed method with unpublished data. Our new approach shed light on internalization of psychological dimensions in LLMs. 

\clearpage
\bibliographystyle{unsrt}  
\bibliography{references}  

\end{document}